\documentclass[10pt,twocolumn,letterpaper]{article}

\usepackage[pagenumbers]{cvpr} 

\usepackage{graphicx}
\usepackage{amsmath}
\usepackage{amssymb}
\usepackage{booktabs}
\usepackage{multirow}
\usepackage{booktabs}
\usepackage{algorithm}
\usepackage{algpseudocode}
\usepackage{mathrsfs}
\usepackage{amsmath}

\usepackage[capitalize]{cleveref}
\crefname{section}{Sec.}{Secs.}
\Crefname{section}{Section}{Sections}
\Crefname{table}{Table}{Tables}
\crefname{table}{Tab.}{Tabs.}

\def\cvprPaperID{9610} 
\def\confName{CVPR}
\def\confYear{2022}

\begin{document}

\title{UTC: A Unified Transformer with Inter-Task Contrastive Learning \\ for Visual Dialog}

\author{Cheng Chen$^{1}$, Yudong Zhu$^{2}$, Zhenshan Tan$^{1}$, Qingrong Cheng$^{1}$, Xin Jiang$^{2}$, Qun Liu$^{2}$, Xiaodong Gu$^{1}$\thanks{This work was done under the guidance of  Yudong Zhu, and the corresponding author is Xiaodong Gu (xdgu@fudan.edu.cn)} \\
$^{1}$Department of Electronic Engineering, Fudan University \\  $^{2}$Huawei Noah’s Ark Lab \\
 {\tt\small \{chengchen19, zstan19, qrcheng17, xdgu\}@fudan.edu.cn}\\
\{ {\tt\small \{Jiang.Xin, qun.liu, zhuyudong3\}@huawei.com}
}
\maketitle

  
\begin{abstract} 
Visual Dialog aims to answer multi-round, interactive questions based on the dialog history and image content. Existing methods either consider answer ranking and generating individually or only weakly capture the relation across the two tasks implicitly by two separate models. The research on a universal framework that jointly learns to rank and generate answers in a single model is seldom explored. In this paper, we propose a contrastive learning-based framework UTC to unify and facilitate both discriminative and generative tasks in visual dialog with a single model. Specifically, considering the inherent limitation of the previous learning paradigm, we devise two inter-task contrastive losses i.e., context contrastive loss and answer contrastive loss to make the discriminative and generative tasks mutually reinforce each other. These two complementary contrastive losses exploit dialog context and target answer as anchor points to provide representation learning signals from different perspectives. We evaluate our proposed UTC on the VisDial v1.0 dataset, where our method outperforms the state-of-the-art on both discriminative and generative tasks and surpasses previous state-of-the-art generative methods by more than 2 absolute points on Recall@1.   
\end{abstract}

\section{Introduction} 

Recently, an increasing amount of attention has been paid to vision and language understanding. Many related tasks in this intersecting field have been designed and introduced for different scenarios, such as Moment Localization with Natural Language \cite{2d_tan,tan2021selective},  Image Captioning \cite{Cornia_2020_CVPR}, Visual Question Answering \cite{vqa-chen2020counterfactual}, and Visual Dialog \cite{vd-lu2017best,vd-niu2019RVA}. Among them, Visual Dialog is designed to interact with humans about an unseen image through continuous communication. In general, there are two types of settings in visual dialog: a discriminative decoder that ranks the predefined answer candidates in the discriminative setting, and a generative decoder that synthesizes the target answer in the generative setting.

Compared with visual question answering, visual dialog not only demands that the agent is able to engage in a question about an image but also requires the agent to fully exploit the clues in previous questions and answers. Thus, the interactions among an answer candidate, a question, a dialog history and an image are the key to produce a correct answer. 

\begin{figure}
\includegraphics[width=0.48\textwidth]{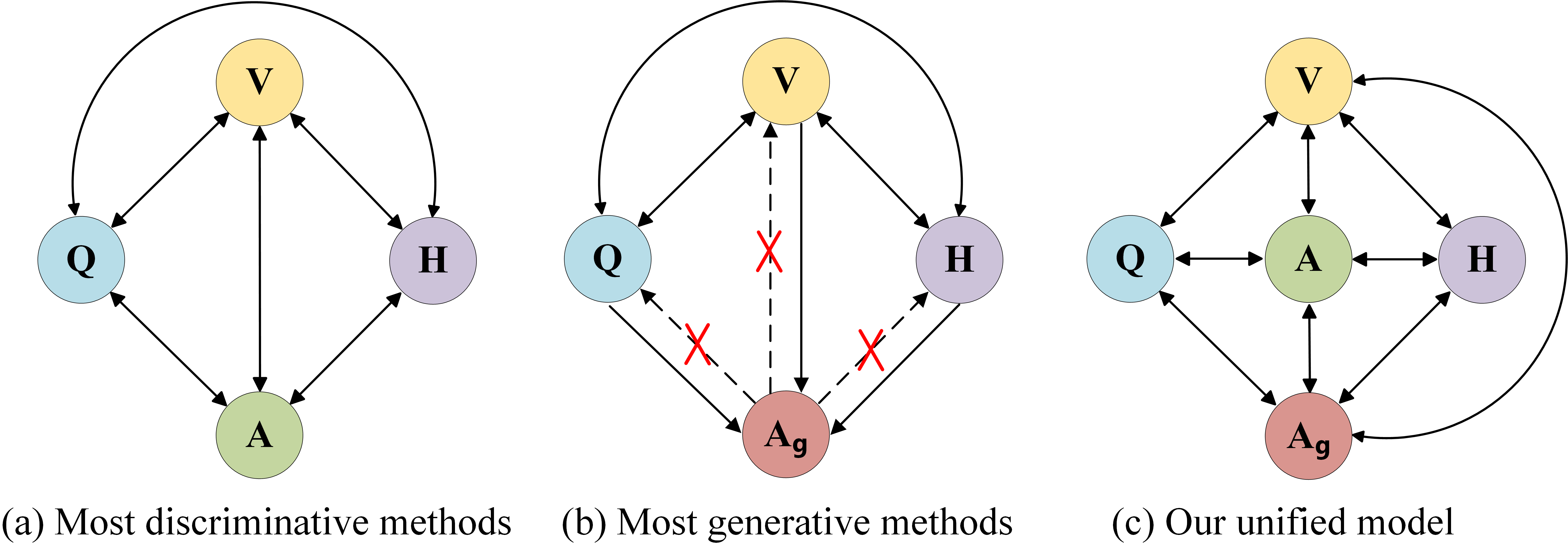}
\caption{Illustration of interaction flow among question (Q), image (V), dialog history (H), answer candidates (A) and generated answer($A_g$).} \label{fig1}
\end{figure}

As shown in Figure~\ref{fig1}(a), most of the current visual dialog models \cite{vdbert,fga} focus on designing various attention mechanisms to capture such interaction in the discriminative setting while training answer ranking and generating tasks individually. Recently, several related works \cite{LTMI,redan} weakly capture the relation across the generative and discriminative tasks by training the entire network using the two decoders simultaneously. Though these models have impressive results, the designing of a unified model to facilitate the training of both answer ranking and generating tasks remains two challenges.

On the one hand, the limitation originates from the inherently different peculiarities of the two tasks. As shown in Figure~\ref{fig1}(a and b), the discriminative setting can capitalize on the unrestricted message passing across answer candidates and multi-modal context. While in the generation task, the models need to autoregressively decode the answer word by word, which makes the message passing from answers to multi-modal context restricted. This raises the first challenge for our unified model: how to fully transfer the rich semantic clues of answer candidates in the discriminative task to answer generation. 

On the other hand, the discriminative setting focuses on the alignment of dialog context and answer, and most of the existing approaches adopt a two-step pipeline. For a given dialog context, answer candidates are first randomly selected from the corresponding answer set. Each answer candidate is then matched with the dialog context to determine whether it is the target answer. This pipeline only considers the relations between one given dialog context and its corresponding answer candidates. It treats each round of dialog separately and neglects to distinguish the dialog context representations. This raises the second challenge: how to capture the dialog contexts relations across other rounds in a dialog and all rounds in other dialogs. Moreover, as aforementioned, the dialog context is enhanced with answer information in the discriminative setting while independent to answer information in the generative setting. It is difficult for previous methods to utilize the enhanced dialog context representations during the training of generative setting since they process each task individually.

In this work, we formulate the interactions of all entities in a unified framework. As shown in Figure~\ref{fig1}(c), the agent ranks and generates the answers with the powerful representation yielded by the interaction of each entity in both settings. Inspired by recent visual and language pretraining, the cross-modal pretrained transformer is employed as the encoder backbone. Instead of employing two types of decoders in prior works, the multi-modal features yielded by the backbone encoder are directly used to rank the answers, and a transformer-based decoder upon the encoder is built to generative the answers.
To tackle the first challenge, we employ the target answer as anchor points to utilize the clues in answer features from the discriminative setting to ease the training of the generative task. Concretely, a contrastive loss is devised to preserve the similarity of the target answer features and generated answer features, while distinguishing other answer option features. It also leads to an elegant view of how to bridge the discrepancy between the discriminative and generative settings and how to exploit the clues in answer candidates efficiently. 

For the second challenge, since the unified model is required to jointly rank and synthesize the target answer based on the dialog context pairs of the discriminative and generative settings respectively, we explicitly promote the model to distinguish the paired dialog context representations from other negative similar dialogs by contrastive learning.
Furthermore, the contrastive learning scheme also enables the generative task to utilize enhanced dialog context representation information from the discriminative task. Our contributions are as follows: 
\begin{itemize}
\item We introduce a unified model for Visual Dialog, which processes all interactions between different entities for both discriminative and generative tasks in a single model. 

\item The target answers and dialog context are employed as anchor points to help facilitate the training of discriminative and generative tasks. Compared with previous methods, two inter-task contrastive losses enable the bidirectional information flow between all answer and dialog context pairs of the two tasks, which significantly eases the training of both tasks.

\item We conduct extensive experiments on VisDial benchmarks to analyze how our model performs on both tasks with various training aspects. The qualitative results indicate that our model obtains reliable improvement on both tasks with inter-task contrastive learning.

\end{itemize}  

\begin{figure*}
\includegraphics[width=\textwidth]{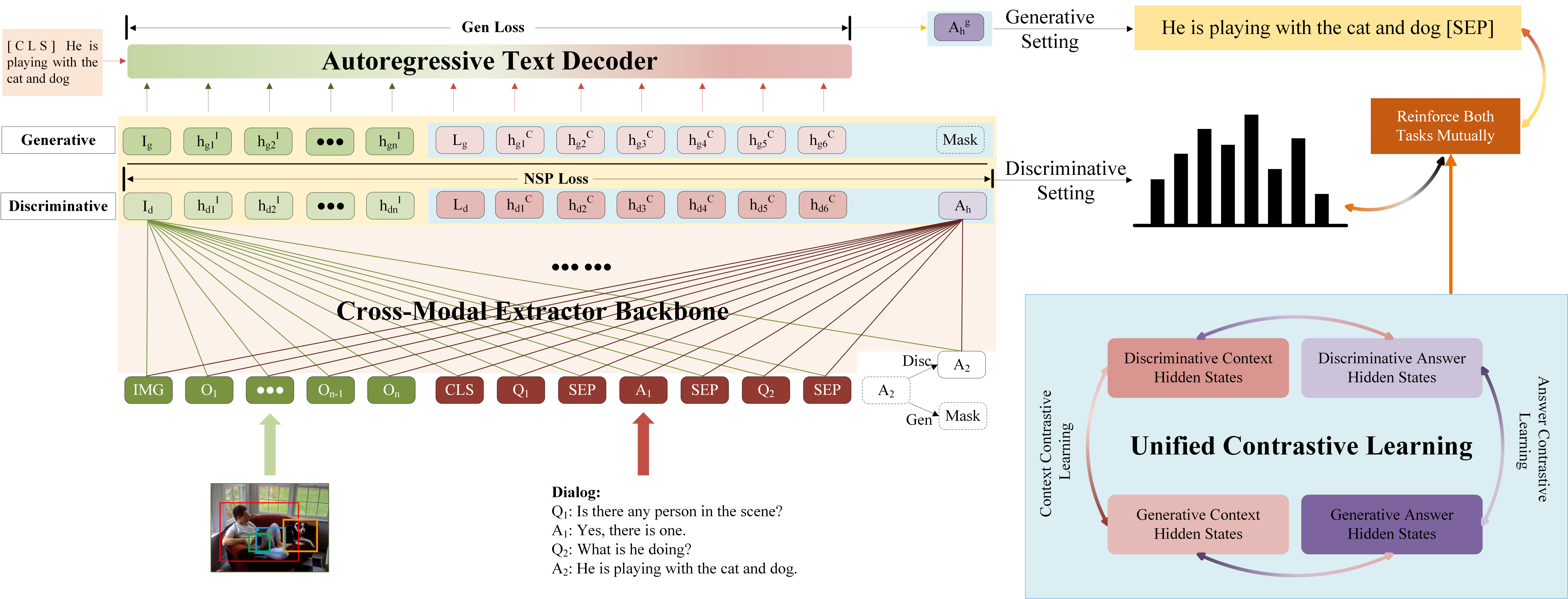}
\caption{The model architecture and contrastive learning paradigm for both discriminative and generative settings of our proposed UTC. For unified contrastive learning, the discriminative context hidden states denote $C_h=\{L_d, h_{d1}^C,..., h_{dt}^C\}$, the discriminative answer hidden states denote $A_h$, the generative context hidden states denote $C_g=\{L_g, h_{g1}^C,..., h_{gt}^C\}$ and the generative answer hidden states denote $A_h^g$.} \label{fig2}
\end{figure*}

\section{Related work} 
\textbf{Vision-Language Task.} There are various vision-language tasks, such as Visual Question Answering \cite{vqa-1},  Image Captioning \cite{ic-you2016image} and Visual Dialog \cite{vd-lu2017best,vd-niu2019RVA}.  
Specifically, for VQA, it is proposed to adopt the attention mechanism to fuse image and text features. Then, based on the fused features, the agent answers the question. 

Recently, pretraining a transformer and its extension has become a popular strategy for multi-modal understanding. 
To be specific, ViLBERT \cite{vilbert} and LXMERT \cite{lxmert} apply two-stream transformers for images and texts respectively. UNITER \cite{uniter} and Oscar \cite{oscar} directly feed the visual features and text embeddings into a unified architecture. These visual-language models are firstly pretrained on external cross-modal datasets then fine-tuned on target tasks, such as visual dialog and image captioning.  

\textbf{Visual Dialog.} Most of the popular approaches employ an encoder-decoder architecture for visual dialog.
The encoder aims at encoding the image and text to fused features, and two separate decoders are employed for ranking and generating respectively. 
Among them, a variety of attention mechanism-based  approaches \cite{dan,redan} are proposed to learn the interactions between the image, the answers, and the dialog history in the discriminative setting. DAN \cite{dan} introduces a dual attention mechanism to reason the key texts in the dialog history. Similarly, FGA \cite{fga} introduces graph attention to model interactions between entities. Besides, granting the dialog model a great ability in grammatical competence also attracts lots of attentions. CorefNMN \cite{Kottur_2018_ECCV} proposes to tackle the visual coreference problem by reasoning over past dialog interactions with previous references. CoAtt \cite{coatt} utilizes adversarial examples to generate more human-like responses.

With the remarkable performance of pretraining transformer, tansformer-based pretrained models with various structures \cite{vdbert, LTMI} are introduced visual dialog. For example, VD-BERT \cite{vdbert} leverages the pre-trained BERT language models for Visual Dialog tasks. LTMI \cite{LTMI} adopts a lightweight transformer to deal with all the interactions between many utilities.

As mentioned above, these methods focus on designing attention mechanisms for answer ranking, which seldom link the two tasks in a unified framework. Different from previous works, the UTC adopts a unified framework to learn vision-language interactions by jointly considering both discriminating and generating tasks, which deeply captures the relations between the two tasks to mutually reinforce each task.
 
\section{Approach} 
As illustrated in Figure~\ref{fig2}, our model consists of three main components: a cross-modal encoder backbone, an autoregressive text decoder and a unified contrastive learning head. The details of these components will be given in the following sections. 

\subsection{Problem Formulation} 
We first formally describe the visual dialog problem. Given a question $Q_t$ grounded on an image $I$ at $t{-th}$ turn, as well as the previous dialog history formulated as $H_t=\{C_{ap};(Q_1;A_1),...,(Q_{t-1};A_{t-1})\}$ (where $C_{ap}$ denotes the caption sentence of the image), our task aims to generate the required answer in generative setting and predict the target answer $A_t$ by ranking a list of 100 answer candidates $\{A_t^1,A_t^2,...,A_t^{100}\}$ in discriminative setting. For simplicity, we define the dialog context $C$ as the dialog history and current question formulated as $C=\{C_{ap};(Q_1;A_1),...,(Q_{t-1};A_{t-1}),Q_t\}$, then the task is to identify whether the answer candidate is correct in discriminative setting and synthesize the required answer in generative setting conditioned on $C$ and $I$. 

\subsection{UTC Architecture} 

\subsubsection{Discriminative Setting}
Following previous work \cite{visdial_bert}, ViLBERT \cite{vilbert} is adopted as the cross-modal extractor backbone network. As shown in Figure~\ref{fig2}, with the pretrained ViLBERT \cite{vilbert}, the contextualized representation for the dialog context, the answer candidate and image can attend to each other bidirectionally, which is convenient for answer prediction in the discriminative setting. 

For a given dialog $D$, the dialog context $C$ and answer candidate $A_t$ will be concatenated as a text sequence: 
{
\begin{small}
\begin{equation} 
D=\{[CLS]C_{ap}[SEP]Q_1[SEP]A_1,...,Q_t[SEP]A_t\}. 
\end{equation}
\end{small}
}
we first extract the image region sequences from Faster R-CNN, and then initialize the [IMG] by mean-pooling of the sequences. 

Following previous works \cite{LTMI,redan}, we first extract the image object bounding box feature sequences $I=\{O_1,...,O_n\}$ from Faster R-CNN\cite{Faster_rcnn}, and then initialize the special learnable token [$IMG$] by mean-pooling of the sequences. Each object feature $O_i$ is a 2048-d Region-of-Interest (RoI) feature and $n$ is the number of the detected objects (fixed to 36 in our setting).

Afterwards, we feed two sequences into ViLBERT and obtain text hidden state $D_h$ and visual hidden state $I_h$ as: 
\begin{small}
\begin{equation} 
D_h, I_h = \textup{ViLBERT}(D,I),    
\end{equation} 
\end{small}
the hidden state $D_h=\{L_d, h_{d1}^C,..., h_{dt}^C, A_h\}$ (here $L_d$, $\{h_{d1}^C,..., h_{dt}^C\}$ and $A_h$ are the hidden states of the [$CLS$] token, the dialog context and the current answer respectively.) and $I_h=\{I_d,h_{d1}^I,...,h_{dn}^I\}$ (here $I_d$ and $\{h_{d1}^I,...,h_{dn}^I\}$ are  the hidden states of the [$IMG$] token and the image object sequence respectively.) are deep interacted cross modal features and will be used to train two tasks. 

As aforementioned, the discriminative task in Visual Dialog is to identify whether the appended answer candidate is correct or not, which is naturally consistent with the pretraining task Next Sentence Prediction (NSP) of ViLBERT. Specifically, the next sentence prediction task in our scenario is trained to predict whether the text input describes the image in ViLBERT. In our visual dialog task, each time we randomly sample an answer candidate and append it to the dialog context, and train the model to distinguish the target answer from other answer candidates by NSP loss. The NSP loss is trained to predict the NSP scores to 1 when the target answer $A_t$ is appended, and 0 when a negative answer $A_n$ is appended. During inference, we rank the answer candidates through the NSP scores. 

\subsubsection{Generative Setting}
As the model is also required to autoregressively generate the answer, we also prepare another text input with the answer mask to make the answer invisible. It is worth mentioning that the mask operation is only performed during training, and the answer is totally removed during inference. Thus the answer information is invisible to the model during inference. The text input is formulated as:
\begin{small}
\begin{equation} 
D_g=\{[CLS]C_{ap}[SEP]Q_1[SEP]A_1,...,Q_t[SEP][MASK]\},   
\end{equation} 
\end{small}
where the answer tokens $A_t$ are fully masked. This masking strategy makes answer information blind to the encoder and only the dialog context is used to autoregressively synthesize the target answer.

To support answer generation, we also feed the text sequence $D_g$ and image $I$ to the ViLBERT backbone. The yielded text hidden state is denoted as $D_h^g=\{L_g, h_{g1}^C,..., h_{gt}^C\}$. 

Given the hidden states $D_h^g$ and $I_h^g$, we additionally devise a cross-modal decoder that learns to reconstruct the masked answer word-by-word. Specifically, we first project $D_h^g$ and $I_h^g$ to a common space as $D_{c}^g$ and $I_{c}^g$, and then perform cross-attention over tokens in $D_{c}^g$ and $I_{c}^g$ for next word prediction. The process can be formulated as: 
\begin{small}
\begin{equation} 
A_h^g=\textup{Decoder}(D_{c}^g,I_{c}^g),   
\end{equation} 
\end{small}
where $A_h^g$ is the generated answer, and the cross-modal decoder is implemented by stacking K transformer-based decoder layers. 

During inference, a [$MASK$] token is recursively appended to the end of the sequence to trigger a one-word prediction and then replace it with the generated token for the next token prediction. The decoding process is terminated when the [$SEP$] token is emitted, and the resulting log-likelihood scores will be used for ranking the answer candidates.

By sharing the cross-modal extractor backbone, our model UTC supports jointly learn two tasks end-to-end. 

\subsection{Unified Contrastive Learning} 
To model the cross-impact and interaction between the discriminative and generative tasks, we enable the task-specific representations to interact with each other via contrastive learning. 

As aforementioned, our model can simultaneously produce the dialog hidden states $D_h$ and $D_h^g$, which originate from discriminative setting and generative setting respectively. We first separate the dialog hidden state $D_h$ in discriminative setting as $C_h$ and $A_h$, here $C_h=\{L_d, h_{d1}^C,..., h_{dt}^C\}$ is dialog context hidden state corresponding to dialog context $C=\{[CLS]C[SEP]Q_1[SEP]A_1,...,Q_t\}$ and $A_h$ is answer candidate hidden state corresponding to the appended answer $A=\{A_t\}$. Note that, as the positional embeddings are added to the sequence, $C_h$ and $A_h$ can be directly divided from text hidden state $D_h$. As the answer tokens are masked in the generative setting, the dialog hidden state $D_h^g$ only contains dialog context information. Thus, we only extract the dialog context hidden state $C_g=\{L_g, h_{g1}^C,..., h_{gt}^C\}$ from $D_h^g$, and the answer hidden state $A_h^g$ is obtained from the output of decoder.
Given the hidden states of dialog context and answer in two tasks, we will next introduce how to utilize these features for unified learning. 

\subsubsection{Answer Contrastive Learning} 
To encourage the decoder to explicitly interact with all rich answer information and optimize the two tasks simultaneously, we leverage the target answer as an anchor and define contrastive losses to transfer useful mutual information between two tasks.

Specifically, when both the answer candidate hidden states $A_h$ and the generated answer hidden states $A_h^g$ are produced, we first divide the answer representations within a batch in the discriminative setting into two parts. More specially, for a given round of dialog context, the target answer representations $A_h^+$ are regarded as the query feature. In our early experiment, the negative samples are only selected from different rounds of a dialog, which shows sub-optimal results. To balance the negative answer samples, we select the negative samples $A_H^-=\{A_{h1}^-,...,A_{hb}^-\}$ (here $b$ is the batch size) from two parts: 1) the corresponding answer candidate set except target answer; 2) all answer options in other dialogs, including all other rounds talking about current image and other irrelevant dialogs associated with different images. 


As the decoder aims to generate target answers, the generated answer hidden states $A_h^g$ need to be semantically close to $A_h^+$. Thus, we utilize the generated answer $A_h^g$ as positive key feature. Note that, the hidden states $A_h^g$, $A_h^+$ and $A_h^-$ are token-level features with different sequence lengths, we first mean-pool the corresponding token features, and obtain the sentence-level features denoted as $A_g^s$, $A_h^{s+}$, and $A_H^{s-}=\{A_{h1}^{s-},...,A_{hb}^{s-}\}$.
Then the answer contrastive loss is thus defined as:
\begin{small}
\begin{equation}
L_{ac}=-log\frac{exp(A_h^{s+}\cdot A_g^s/\tau)}{\sum_{i=0}^{b-1}exp(A_h^{s+} \cdot A_{hi}^{s-}/\tau)},
\end{equation}
\end{small}
where the dot product denotes the cosine similarity score and $\tau$ is a temperature parameter. 

\subsubsection{Context Contrastive Learning} 
UTC can produce dialog context state $C_h$ and $C_g$ of discriminative and generative tasks respectively in an end-to-end manner. It is required to jointly identify the target answer from answer set and synthesize the answer based on $C_h=\{L_d, h_{d1}^C,..., h_{dt}^C\}$ and $C_g=\{L_g, h_{g1}^C,..., h_{gt}^C\}$ respectively. Thus, the hidden state $C_h$ and $C_g$ of a given round of dialog should be semantically close to each other.  

During training, we encourage the dialog context representations $C_h$ and $C_g$ from one round of dialog to be more semantically close. Besides, $C_h$ should be distinct from the context representations of other rounds and all other dialogs.

For a given round of dialog, the dialog context representations in the discriminative setting are regarded as query feature $C_h^+$.
The positive key features are denoted as $C_g$ which originates from the generative task. In the common space, we aim to simultaneously minimize the distance between $C_h^+$ and $C_g$ while maximize the distances between $C_h^+$ and negative key feature set $C_h^-=\{C^-_{h1},...,C^-_{hb}\}$. $C_h^-$ comes from all other rounds and other dialogs within a batch. Similar to answer features, we mean-pool the context token features $C_h^+$, $C_g$ and $C_h^-$ to obtain the corresponding sentence-level features $C_h^{s+}$, $C_g^s$ and $C_h^{s-}=\{C^{s-}_{h1},...,C^{s-}_{hb}\}$.  

We thus formulate the context contrastive loss as:
\begin{small}
\begin{equation}
L_{cc}=-log\frac{exp(C_h^{s+} \cdot C_g^s/\tau)}{\sum_{i=0}^{b-1}exp(C_h^{s+} \cdot C^{s-}_{hi}/\tau)},
\end{equation}
\end{small}
in which $\tau$ is a temperature parameter and the dot product denotes the cosine similarity score. 

\subsection{Training Objectives} 
During the training of UTC, We use two visually grounded training objectives masked language modeling (MLM) and next sentence prediction (NSP) to supervise the cross-modal extractor backbone ViLBERT. 

Similar to MLM in BERT, 10\% tokens in text input and 15\% tokens in visual input are randomly masked out and replaced with a special token [$MASK$]. The model is required to recover them based on the surrounding tokens $D_{\backslash m}$ and the cross-modal clues $I_{\backslash m}$: 

\begin{small}
\begin{equation} 
L_{mlm}=-E_{(D,I)\sim T}\textup{log}P(W_m|D_{\backslash m},I_{\backslash m}), 
\end{equation} 
\end{small}
where $W_m$ is the masked tokens and $T$ refers to the training set. 
The NSP loss aims to identify whether the appended answer candidate is correct or not, which is implemented based on jointly understanding the text and image as: 

\begin{small}
\begin{equation} 
L_{nsp}=-E_{(D,I)\sim T}\textup{log}P(y|N(D,I)), 
\end{equation} 
\end{small}
where $y\in\{0,1\}$ serves as the supervision label, and $N(\dot)$ is the binary answer prediction head to predict the probability based on the dot product of [$CLS$] token representation and [$IMG$] token representation. 

For the generative setting, the decoder is required to reconstruct the sequential answer tokens depending on all the dialog context and input image. The loss is defined as a maximum log-likelihood loss: 

\begin{small}
\begin{equation} 
L_g=-E_{(D,I)\sim T}\textup{log}P(A|D_{\backslash A},I),
\end{equation} 
\end{small}
We formulate the final loss for our unified contrastive training method as: 

\begin{small}
\begin{equation} 
L_{utc}=L_{mlm}+L_{nsp}+\alpha L_{g}+L_{ac}+L_{cc}, 
\end{equation} 
\end{small}
where $\alpha=0.05$ is the weighting parameter. 
\section{Experiment} 

\subsection{Dataset} 

We evaluate the proposed UTC approach on the VisDial v1.0 dataset. It has 123,287, 2,064, and 8,000 images for training, validation and testing, respectively. Each image is associated with a caption sentence and 10 question-answer pairs. For each round of question-answer pairs, 100 answer candidates are given. The validation split and part of the train split (2,000 images) are provided with dense annotations (i.e., relevance scores) for all candidate answers.

\begin{table}\setlength{\tabcolsep}{1.0pt}
\fontsize{9}{10}\selectfont
\centering
\caption{Performance comparisons of discriminative setting on the val split of VisDial v1.0 dataset. The top 2 results are highlighted by $\mathbf {bold}$ and \underline{underline} respectively. The remaining tables follow the same notations. }\label{tab1}
\begin{tabular}{lccccccc}
\toprule
Methods&R@1$\uparrow$ &R@5$\uparrow$ & R@10$\uparrow$& NDCG$\uparrow$& MRR$\uparrow$ &Mean$\downarrow$\\ 
\midrule
MN & 46.09 & 78.14 & 88.05 & 55.13 & 60.42 & 4.63 \\
CoAtt & 48.86 & 80.41 & 89.83 & 57.72 & 62.91 & 4.21 \\
VisDial-BERT & 53.42 & \underline{84.41} & \underline{92.62} & 60.96 & 67.17 & \underline{3.41} \\
HCIAE & 48.94 & 80.50 & 89.66 & 57.75 & 62.96 & 4.24 \\
ReDAN & 50.60 & 81.39 & 90.26 & 59.32 & 64.21 & 4.05 \\
LTMI & 48.94 & 78.65 & 87.88 & 62.72 & 62.32 & 4.86 \\
VDBERT & \underline{54.02} & 83.96 & 92.33 & \underline{63.22} & \underline{67.44} & 3.53 \\

\midrule
$\mathbf {UTC}$ & $\mathbf {55.48}$ & $\mathbf {85.38}$ & $\mathbf {93.20}$ & $\mathbf {63.22}$ & $\mathbf {68.58}$ & $\mathbf {3.28}$ \\
\midrule
\end{tabular}
\end{table}

\begin{table}\setlength{\tabcolsep}{1.1pt}
\fontsize{10}{10}\selectfont
\centering
\caption{Performance comparisons of generative setting on the val split of VisDial v1.0 dataset. The top 1 results are highlighted by $\mathbf {bold}$. }\label{tab2}
\begin{tabular}{lccccccc}
\toprule
Methods&R@1$\uparrow$ &R@5$\uparrow$ & R@10$\uparrow$& NDCG$\uparrow$& MRR$\uparrow$ &Mean$\downarrow$\\ 
\midrule
MN & 38.01 & 57.49 & 64.08 & 56.99 & 47.83 & 18.76 \\
CoAtt & 40.09 & 59.37 & 65.92 & 59.24 & 49.64 & 17.86 \\
HCIAE & 39.72 & 58.23 & 64.73 & 59.70 & 49.07 & 18.43 \\
LTMI & \underline{40.44} & \underline{61.61} & $\mathbf{69.71}$ & \underline{63.58} & \underline{50.74} & $\mathbf{14.93}$ \\
ReDAN & 40.27 & 59.93 & 66.78 & 60.47 & 50.02 & 17.40 \\
\midrule
$\mathbf {UTC}$ & $\mathbf {42.56}$ & $\mathbf {62.40}$ & $\underline{69.51}$ & $\mathbf {63.86}$ & $\mathbf {52.22}$ & $\underline{15.67}$ \\
\midrule
\end{tabular}
\end{table}

\subsection{Evaluation Metric} 

Following previous works \cite{LTMI,redan,fga}, the ranking metrics like Recall@K (K=1, 5, 10), Mean Reciprocal 
Rank (MRR) and Mean Rank are adopted. Since the 2018 VisDial challenge releases the dense annotations of each answer option's relevance degree, Normalized Discounted Cumulative Gain (NDCG) that penalizes the low ranked answer options with high relevance is also used. 

\subsection{Implementation Details} 
We use ViLBERT BASE as the backbone, which has 12 layers of transformer blocks with each block having a hidden state size of 768 and 12 attention heads. The decoder consists of 12 layers of transformer blocks, each block has a hidden size of 1024 and 16 attention heads. For a fair comparison, the backbone ViLBERT is initialized with the weights that are pretrained on the Visual Question Answering dataset \cite{VQA} like previous works. And the decoder is trained from scratch. 
The max text sequence length is 256. We train the UTC on 8 V100 GPUs with a batch size of 120 for 20 epochs. The Adam optimizer with initial learning rates of 2e-4 is adopted. A linear decay learning rate schedule with warmup is employed to train the model.

\subsection{Comparison to State-of-the-Art Methods} 
We compare our method with recently published methods on the VisDial v0.9 and VisDial v1.0 datasets, including LF \cite{das-visual}, MN \cite{das-visual}, MCA \cite{mca}, MN-Att (with attention) \cite{das-visual}, CoAtt \cite{coatt}, FGA \cite{fga}, RvA \cite{vd-niu2019RVA}, DAN \cite{dan}, ReDAN \cite{redan}, GNN \cite{gnn}, HCIAE \cite{vd-lu2017best}, LTMI \cite{LTMI}, VDBERT \cite{vdbert}, VisDial-BERT \cite{visdial_bert}, and Synergistic \cite{Synergistic}. 

\subsubsection{Results on the VisDial v1.0 val} 
\begin{table}\setlength{\tabcolsep}{1.1pt}
\fontsize{10}{10}\selectfont
\centering
\caption{Performance comparisons on the test split of VisDial v1.0 dataset. The results are reported by the test server. * denotes fine-tuning on dense annotations. }\label{tab3}
\begin{tabular}{lccccccc}
\toprule
Methods&R@1$\uparrow$ &R@5$\uparrow$ & R@10$\uparrow$& NDCG$\uparrow$& MRR$\uparrow$ &Mean$\downarrow$\\ 
\midrule
LF & 40.95 & 72.45 & 82.23 & 45.31 & 55.42 & 5.95 \\

MN & 40.98 & 72.30 & 83.30 & 47.50 & 55.49 & 5.92 \\

MN-Att & 42.42 & 74.00 & 84.35 & 49.58 & 56.90 & 5.59 \\

FGA & 52.75 & 82.92 & 91.07 & 56.90 & 66.20 & 3.80 \\
DAN & 49.63 & 79.75 & 89.35 & 57.59 & 63.20 & 4.30 \\
Synergistic & 47.90 & 80.43 & 89.95 & 57.32 & 62.20 & 4.17 \\
LTMI & 50.20 & 80.68 & 90.35 & 59.03 & 64.08 & 4.05 \\
VisDial-BERT & \underline{53.85} & \underline{84.68} & $\mathbf{93.25}$ & \underline{63.87} & \underline{67.50} & \underline{3.32} \\
VDBERT & 51.63 & 82.23 & 90.68 & 59.96 & 65.44 & 3.90 \\
RvA & 49.03 & 80.40 & 89.83 & 55.59 & 63.03 & 4.18 \\
GNN & 47.33 & 77.98 & 87.83 & 52.82 & 61.37 & 4.57\\
VDBERT* & 33.15 & 61.58 & 77.15 & $\mathit {74.54}$ & 50.74 & 7.18 \\
MCA* & 20.67 & 56.67 & 72.12 & 72.47 & 37.68 & 8.89 \\
\midrule
$\mathbf {UTC}$ & 52.25 & 83.55 & 92.23& 62.65 & 66.27& 3.48 \\
$\mathbf{UTC_{vqa+cc}}$ & $\mathbf {55.73}$ & $\mathbf {84.93}$ & \underline{93.08} & $\mathbf {64.60}$ & $\mathbf {68.70}$ & $\mathbf {3.32}$ \\
UTC* & 37.12 & 63.98 & 79.88 & \underline{74.32} & 50.24 & 6.48\\
\midrule
\end{tabular}
\end{table}

\begin{table*}[t]\setlength{\tabcolsep}{3pt}
    \centering
    \caption{Ablation studies on the val v1.0 split
of VisDial dataset.}
    \begin{tabular}{lcccccc|cccccc}
        \toprule
        \multirow{2}*{\bfseries Methods}  &
        \multicolumn{6}{c}{\bfseries Discriminative} & \multicolumn{6}{c}{\bfseries Generative}\\

        \cmidrule{2-13} &R@1&R@5&R@10&NDCG$\uparrow$&MRR$\uparrow$&Mean$\downarrow$&R@1&R@5&R@10&NDCG$\uparrow$&MRR$\uparrow$&Mean$\downarrow$\\

        \midrule
        UTC$_{individual}$ &53.94&84.10&92.17&61.20&67.29&3.48&41.39&59.85&66.33&61.04& 50.61 &17.70 \\
        UTC$_{elementary}$
        &54.39&84.36&92.35&61.47&67.69&3.44&41.75&60.34&66.76&61.72 & 50.92 & 17.35 \\
     
        UTC$_{w/o-L_{cc}}$
        &54.55&84.95&92.95&62.02&67.97&3.35&\underline{42.52}&\underline{62.01}&\underline{69.14}&\underline{63.15} & \underline{52.02} & \underline{15.71} \\
        UTC$_{w/o-L_{ac}}$
        &\underline{55.24}&\underline{85.05}&\underline{93.16}&\underline{62.91}&\underline{68.48}&\underline{3.29}&42.09&61.66&68.28&62.65 & 51.67 & 15.88 \\ 
        UTC
         & $\mathbf {55.48}$ & $\mathbf {85.38}$ & $\mathbf {93.20}$ & $\mathbf {63.22}$ & $\mathbf {68.58}$ & $\mathbf {3.28}$ & $\mathbf {42.56}$ & $\mathbf {62.40}$ & $\mathbf {69.51}$ & $\mathbf {63.86}$ & $\mathbf {52.22}$ & $\mathbf {15.67}$  \\
        \midrule
    \end{tabular}
    \centering
    \label{tab4}
\end{table*}

We first compare our model with state-of-the-art approaches on the val v1.0 split. The results of discriminative setting and generative setting are shown in Table~\ref{tab1} and Table~\ref{tab2} respectively. Since VisDial-BERT only supports the discriminative setting and VDBERT doesn't report the performance of the generative setting on the VisDial v1.0 val, we compare the performances of the discriminative setting with them. 
The results show that UTC outperforms other competitors in various scenarios on both tasks across different criteria. In all cases, UTC ranks the first or the second. In more detail, compared with other related methods, we obtain several observations.  

First, we compare our model with pretraining methods (i.e. VDBERT and VisDial-BERT), which achieve state-of-the-art on the discriminative setting. VisDial-BERT is pretrained not only on the VQA dataset but also on external large-scale vision-language datasets like Conceptual Captions. Since they didn't provide their pre-training weights on two pretraining dataset and only released their best model fine-tuned on VisDial v1.0 dataset.  We first compare its results pretrained on the VQA dataset for fair a comparison in Table~\ref{tab1}. Due to the powerful cross-modal representation ability of pretraining, the pretrained transformer-based approaches perform better than the traditional approaches on the discriminative setting, however, inferior to our proposed UTC method. Moreover, we implement our unified framework on VisDial-BERT since it only support discriminative task. Then we further train their best fine-tuned model with our contrastive losses and compare with it on test spilt. The results in Table~\ref{tab3} (UTC$_{vqa+cc}$) show our contrastive learning paradigm can further gain improvements over it, which demonstrate the effectiveness of our learning paradigm.


Furthermore, we compare UTC with attention-based method LTMI, which achieves the state-of-the-art on the generative setting. It utilizes a transformer-based structure to deal with all the interactions in visual dialog while weakly captures the relations between two tasks. From Table~\ref{tab2}, we can see that our UTC outperforms LTMI by more than 2 points in terms of Recall@1 on the generative setting, while slightly inferior to it at Recall@10 and mean rank. It further validates that UTC can simultaneously predict and generate answers accurately. 

\subsubsection{Results on the VisDial v1.0 test} 

We next report the comparison results on the test-standard v1.0 split. The results are shown in Table~\ref{tab3}. As the ground-truth answers and the dense annotations of the test v1.0 split are not publicly available, we upload our predicted results to the task organizers' server for evaluation. The results show that our single-model UTC significantly outperforms other single-model methods across various metrics. Compared with VD-BERT, the current state-of-the-art method on the VisDial v1.0 dataset, our model pretrained only on VQA improves NDCG from 59.96 to 62.65.

Moreover, we follow previous works \cite{LTMI, vdbert} to further fine-tune UTC on the available dense annotations, where a cross-entropy loss with soft labels (i.e. relevance scores) is minimized. For each round of dialog, the answers and dialog contexts pairs with relevance scores higher than zero are removed from the negative key set when calculating two contrastive losses.
In this case, our model achieves superior results than other single-model methods. Similar to previous works \cite{LTMI,vdbert}, the accuracy values measured by NDCG and other metrics show a trade-off relation. It can be observed that finetuning on dense annotations significantly increases NDCG while hurts other metrics.

\begin{figure}
\includegraphics[width=0.48\textwidth]{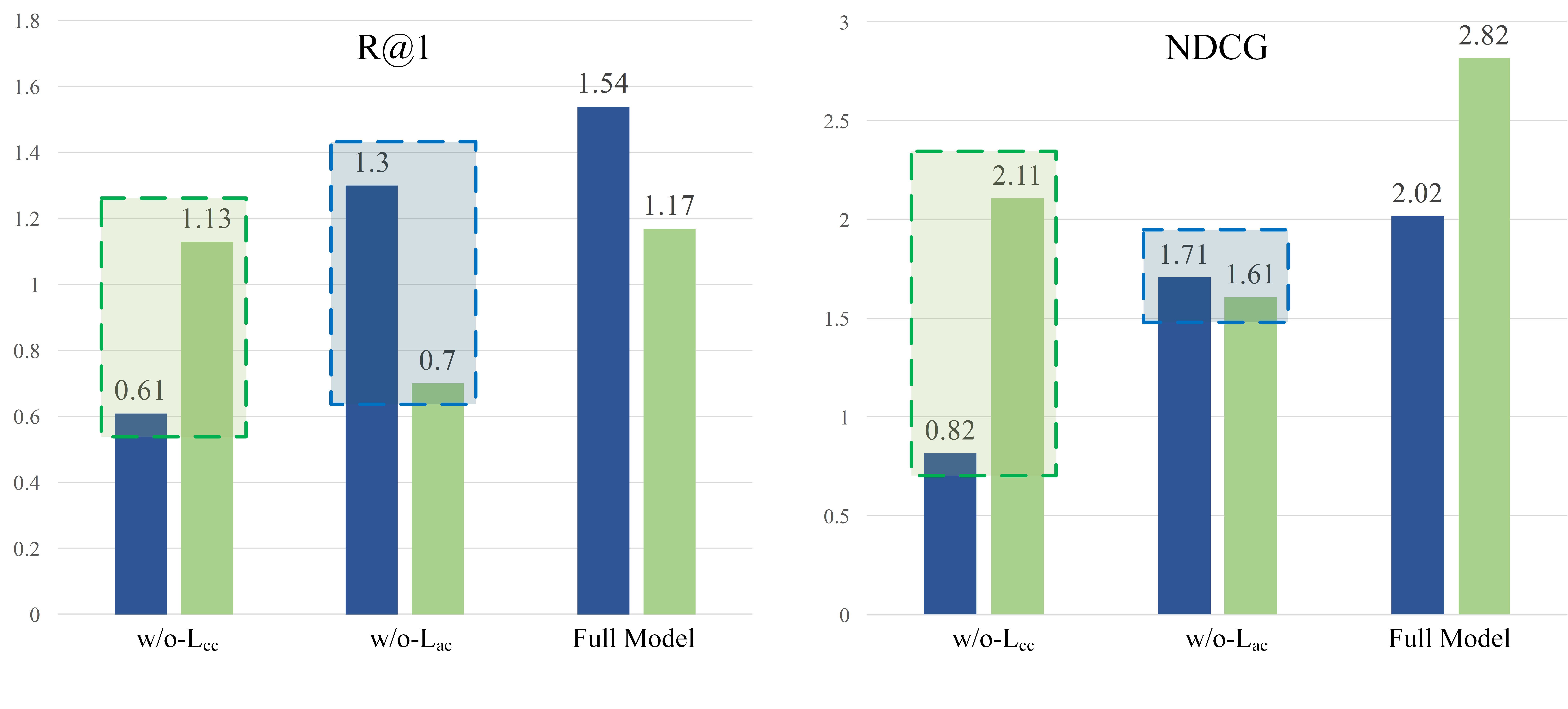}
\caption{Illustration of the metric improvements, where the blue and green histograms represent discriminative and generative tasks respectively.} \label{fig3}
\end{figure}
\subsection{Ablation Studies} 
\textbf{Baselines.} In this section, we perform ablation studies to evaluate the effects of different training settings. The results are shown in Table~\ref{tab4}. UTC$_{individual}$ in the first row stands for training two tasks individually. UTC$_{elementary}$ in the second row stands for training two tasks by simply minimizing the sum of two task losses without contrastive learning. Comparing UTC$_{individual}$ and UTC$_{elementary}$, it can be observed that the training of the generative task brings improvements to the ranking task. The main characteristic of UTC is the unified contrastive loss, which combines all dialog context and answers features from two tasks to learn more valuable clues. 

\begin{figure*}[h]
\includegraphics[width=\textwidth]{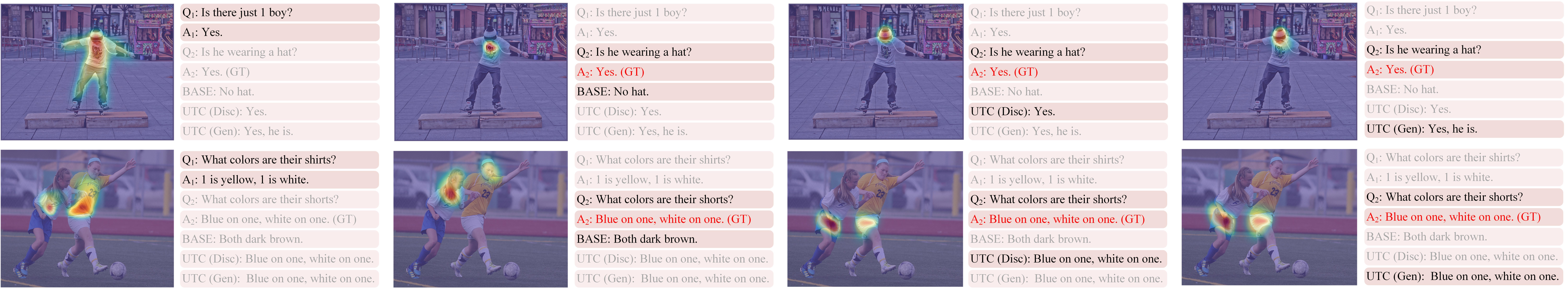}
\caption{The effects of contrastive learning on two tasks in UTC, the first column is the correct attention weights yielded from the discriminative setting and the other three columns are the attention weights corresponding to the base model, discriminative setting and generative setting respectively.} \label{fig4}
\end{figure*}

\textbf{Does combining two losses help?} To study the impact of the contrastive loss alone, we first remove context contrastive loss and train our UTC with only answer contrastive loss. The results are reported in the third row UTC$_{w/o-L_{cc}}$. Comparing to the base model with contrastive loss (UTC$_{elementary}$), UTC$_{w/o-L_{cc}}$ gets better performance across various metrics in both discriminative and generative tasks. For simplicity, we illustrate the absolute improvement values of NDCG and R@1 relative to the UTC$_{individual}$ in Figure~\ref{fig3}. 
It can be observed in both Figure~\ref{fig3} and Table~\ref{tab4} that the performance increases in the generative setting are more significant than that in the discriminative setting with only answer contrastive loss. The underlying reason is that the discriminative task densely samples answer candidates to calculate the NSP loss, which has a similar impact of answer contrastive learning. However, the generative task can not see the answer information without contrastive learning. Thus, the performance of UTC increases significantly in the generative task with answer contrastive learning.

We further compare the effectiveness of context contrastive learning, and the results are shown in the fourth row UTC$_{w/o-L_{ac}}$. We observe that context contrastive learning also brings improvements to both tasks. This phenomenon further proves our hypothesis that unified learning of two tasks enables networks to exploit useful information across different tasks. In contrast to answer contrastive learning, context contrastive learning brings more improvements to the discriminative task. It is because that treating each dialog context individually in previous methods is not sufficient to learn discriminative dialog context representation.
On the other hand, the answer generation results depend on not only the dialog context representation but also the decoding process. Hence, UTC gains large improvements with context contrastive learning on the discriminative task compared to the generative task.

The results of our full model with two contrastive losses are shown in the fifth row of Table~\ref{tab4} and last column of Figure~\ref{fig3}. Interestingly, the two losses show complementary properties. With two complementary inter-task losses providing representation learning signals from different perspectives, our full model achieves the best performance on different criteria of both tasks.

\subsection{Qualitative Result} 
To interpret the unified contrastive learning our UTC, we visualize the attention weights of the self attention layers in Figure~\ref{fig4}. In these examples, consistent with the Visual Dialog task definition, we show the top-1 prediction. The last layers of the encoder and decoder corresponding to discriminative and generative settings are used here. We compare UTC against the elementary unified model trained without two contrastive losses. It can be observed that the discriminative and generative settings of UTC tends to focus on the similar regions in the image. In the first example, the hat are highlighted at both setting, where it is the key to answer the question.

In most cases, training with two contrastive losses can produce more accurate results. For example, in the second row, the questions mention multiple objects. It is very difficult for the model to predict the target answer without proper reference to the visual information. As our model exploits contrastive learning to learn rich information from all answer candidates, it correctly grounds the entities like shirt and short in the image, thus performs better than the baseline on both discriminative and generative tasks.
However, as we push the dialog context and answer features of both settings to be close, we also find that our model may be simultaneously confused on both tasks when it is uncertain to predict the target answer. We aim to study it in the future.

\section{Conclusion} 
In this paper, we propose a unified Transformer model UTC that exploits the dialog context and target answer as anchor points for jointly training discriminative and generative tasks. UTC is capable of modeling all the interactions to rank and generate answers seamlessly in an end-to-end manner. Moreover, two complementary contrastive losses are defined to facilitate the training of two tasks. 
Experiments on Visual Dialog benchmarks show the effectiveness of the proposed model, and more extensive ablation studies further confirm the correlation between two tasks and reveal that modeling the relations explicitly by inter-task contrastive learning can improve their performances. 

Our UTC can be formalized as a unified framework of discriminative and generative tasks. It is easy to transfer the contrastive learning scheme to other tasks. In the future, we will explore the application of our
framework to more scenarios.

\textbf{Acknowledgements.} This work was supported in part by National Natural Science Foundation of China under grant 62176062. Thanks Yi Wu for her kindly support.
{\small
\bibliographystyle{ieee_fullname}
\bibliography{egbib}

\begin{thebibliography}{10}\itemsep=-1pt

\bibitem{mca}
Shubham Agarwal, Trung Bui, Joon-Young Lee, Ioannis Konstas, and Verena Rieser.
\newblock History for visual dialog: Do we really need it?
\newblock In {\em Proceedings of the 58th Annual Meeting of the Association for
  Computational Linguistics}, pages 8182--8197, Online, July 2020. Association
  for Computational Linguistics.

\bibitem{vqa-1}
Stanislaw Antol, Aishwarya Agrawal, Jiasen Lu, Margaret Mitchell, Dhruv Batra,
  C~Lawrence Zitnick, and Devi Parikh.
\newblock Vqa: Visual question answering.
\newblock In {\em Proceedings of the IEEE international conference on computer
  vision}, pages 2425--2433, 2015.

\bibitem{VQA}
Stanislaw Antol, Aishwarya Agrawal, Jiasen Lu, Margaret Mitchell, Dhruv Batra,
  C.~Lawrence Zitnick, and Devi Parikh.
\newblock {VQA}: {V}isual {Q}uestion {A}nswering.
\newblock In {\em International Conference on Computer Vision (ICCV)}, 2015.

\bibitem{vqa-chen2020counterfactual}
Long Chen, Xin Yan, Jun Xiao, Hanwang Zhang, Shiliang Pu, and Yueting Zhuang.
\newblock Counterfactual samples synthesizing for robust visual question
  answering.
\newblock In {\em Proceedings of the IEEE/CVF Conference on Computer Vision and
  Pattern Recognition}, pages 10800--10809, 2020.

\bibitem{uniter}
Yen-Chun Chen, Linjie Li, Licheng Yu, Ahmed El~Kholy, Faisal Ahmed, Zhe Gan, Yu
  Cheng, and Jingjing Liu.
\newblock Uniter: Universal image-text representation learning.
\newblock In {\em European conference on computer vision}, pages 104--120.
  Springer, 2020.

\bibitem{Cornia_2020_CVPR}
Marcella Cornia, Matteo Stefanini, Lorenzo Baraldi, and Rita Cucchiara.
\newblock Meshed-memory transformer for image captioning.
\newblock In {\em Proceedings of the IEEE/CVF Conference on Computer Vision and
  Pattern Recognition (CVPR)}, June 2020.

\bibitem{das-visual}
Abhishek Das, Satwik Kottur, Khushi Gupta, Avi Singh, Deshraj Yadav,
  Jos{\'e}~MF Moura, Devi Parikh, and Dhruv Batra.
\newblock Visual dialog.
\newblock In {\em Proceedings of the IEEE Conference on Computer Vision and
  Pattern Recognition}, pages 326--335, 2017.

\bibitem{redan}
Zhe Gan, Yu Cheng, Ahmed~El Kholy, Linjie Li, Jingjing Liu, and Jianfeng Gao.
\newblock Multi-step reasoning via recurrent dual attention for visual dialog.
\newblock {\em arXiv preprint arXiv:1902.00579}, 2019.

\bibitem{Synergistic}
Dalu Guo, Chang Xu, and Dacheng Tao.
\newblock Image-question-answer synergistic network for visual dialog.
\newblock pages 10426--10435, 06 2019.

\bibitem{dan}
Gi-Cheon Kang, Jaeseo Lim, and Byoung-Tak Zhang.
\newblock Dual attention networks for visual reference resolution in visual
  dialog.
\newblock {\em arXiv preprint arXiv:1902.09368}, 2019.

\bibitem{Kottur_2018_ECCV}
Satwik Kottur, Jos\'e M.~F. Moura, Devi Parikh, Dhruv Batra, and Marcus
  Rohrbach.
\newblock Visual coreference resolution in visual dialog using neural module
  networks.
\newblock In {\em The European Conference on Computer Vision (ECCV)}, September
  2018.

\bibitem{oscar}
Xiujun Li, Xi Yin, Chunyuan Li, Pengchuan Zhang, Xiaowei Hu, Lei Zhang, Lijuan
  Wang, Houdong Hu, Li Dong, Furu Wei, et~al.
\newblock Oscar: Object-semantics aligned pre-training for vision-language
  tasks.
\newblock In {\em European Conference on Computer Vision}, pages 121--137.
  Springer, 2020.

\bibitem{vilbert}
Jiasen Lu, Dhruv Batra, Devi Parikh, and Stefan Lee.
\newblock Vilbert: Pretraining task-agnostic visiolinguistic representations
  for vision-and-language tasks.
\newblock {\em arXiv preprint arXiv:1908.02265}, 2019.

\bibitem{vd-lu2017best}
Jiasen Lu, Anitha Kannan, Jianwei Yang, Devi Parikh, and Dhruv Batra.
\newblock Best of both worlds: Transferring knowledge from discriminative
  learning to a generative visual dialog model.
\newblock In {\em NIPS}, 2017.

\bibitem{visdial_bert}
Vishvak Murahari, Dhruv Batra, Devi Parikh, and Abhishek Das.
\newblock Large-scale pretraining for visual dialog: A simple state-of-the-art
  baseline.
\newblock {\em arXiv preprint arXiv:1912.02379}, 2019.

\bibitem{LTMI}
Van-Quang Nguyen, Masanori Suganuma, and Takayuki Okatani.
\newblock Efficient attention mechanism for visual dialog that can handle all
  the interactions between multiple inputs.
\newblock In {\em Computer Vision--ECCV 2020: 16th European Conference,
  Glasgow, UK, August 23--28, 2020, Proceedings, Part XXIV 16}, pages 223--240.
  Springer, 2020.

\bibitem{vd-niu2019RVA}
Yulei Niu, Hanwang Zhang, Manli Zhang, Jianhong Zhang, Zhiwu Lu, and Ji-Rong
  Wen.
\newblock Recursive visual attention in visual dialog.
\newblock In {\em Proceedings of the IEEE/CVF Conference on Computer Vision and
  Pattern Recognition}, pages 6679--6688, 2019.

\bibitem{Faster_rcnn}
Shaoqing Ren, Kaiming He, Ross Girshick, and Jian Sun.
\newblock Faster r-cnn: Towards real-time object detection with region proposal
  networks.
\newblock {\em IEEE Transactions on Pattern Analysis and Machine Intelligence},
  39(6):1137--1149, 2017.

\bibitem{fga}
Idan Schwartz, Seunghak Yu, Tamir Hazan, and Alexander~G Schwing.
\newblock Factor graph attention.
\newblock In {\em Proceedings of the IEEE/CVF Conference on Computer Vision and
  Pattern Recognition}, pages 2039--2048, 2019.

\bibitem{lxmert}
Hao Tan and Mohit Bansal.
\newblock Lxmert: Learning cross-modality encoder representations from
  transformers.
\newblock {\em arXiv preprint arXiv:1908.07490}, 2019.

\bibitem{tan2021selective}
Yi Tan, Yanbin Hao, Xiangnan He, Yinwei Wei, and Xun Yang.
\newblock Selective dependency aggregation for action classification.
\newblock In {\em Proceedings of the 29th ACM International Conference on
  Multimedia}, pages 592--601, 2021.

\bibitem{vdbert}
Yue Wang, Shafiq Joty, Michael~R Lyu, Irwin King, Caiming Xiong, and Steven~CH
  Hoi.
\newblock Vd-bert: A unified vision and dialog transformer with bert.
\newblock {\em arXiv preprint arXiv:2004.13278}, 2020.

\bibitem{ic-you2016image}
Quanzeng You, Hailin Jin, Zhaowen Wang, Chen Fang, and Jiebo Luo.
\newblock Image captioning with semantic attention.
\newblock In {\em Proceedings of the IEEE conference on computer vision and
  pattern recognition}, pages 4651--4659, 2016.

\bibitem{coatt}
Zhou Yu, Jun Yu, Jianping Fan, and Dacheng Tao.
\newblock Multi-modal factorized bilinear pooling with co-attention learning
  for visual question answering.
\newblock {\em IEEE International Conference on Computer Vision (ICCV)}, pages
  1839--1848, 2017.

\bibitem{2d_tan}
Songyang Zhang, Houwen Peng, Jianlong Fu, and Jiebo Luo.
\newblock Learning 2d temporal adjacent networks for moment localization with
  natural language.
\newblock {\em Proceedings of the AAAI Conference on Artificial Intelligence},
  34(07):12870--12877, Apr. 2020.

\bibitem{gnn}
Zilong Zheng, Wenguan Wang, Siyuan Qi, and Song-Chun Zhu.
\newblock Reasoning visual dialogs with structural and partial observations.
\newblock In {\em Proceedings of the IEEE/CVF Conference on Computer Vision and
  Pattern Recognition}, pages 6669--6678, 2019.

\end{thebibliography}
}

\end{document}